# Microscopic Nuclei Classification, Segmentation and Detection with improved Deep Convolutional Neural Network (DCNN) Approaches

Md Zahangir Alom, Chris Yakopcic, Tarek M. Taha, and Vijayan K. Asari
Department of Electrical and Computer Engineering, University of Dayton, OH, USA
Emails: {alomm1, cyakopcic1, ttaha1, vasari1}@udayton.edu

Due to cellular heterogeneity, cell nuclei classification, segmentation, and detection from pathological images are challenging tasks. In the last few years, Deep Convolutional Neural Networks (DCNN) approaches have been shown state-of-the-art (SOTA) performance on histopathological imaging in different studies. In this work, we have proposed different advanced DCNN models and evaluated for nuclei classification, segmentation, and detection. First, the Densely Connected Recurrent Convolutional Network (DCRN) model is used for nuclei classification. Second, Recurrent Residual U-Net (R2U-Net) is applied for nuclei segmentation. Third, the R2U-Net regression model which is named UD-Net is used for nuclei detection from pathological images. The experiments are conducted with different datasets including Routine Colon Cancer(RCC) classification and detection dataset, and Nuclei Segmentation Challenge 2018 dataset. The experimental results show that the proposed DCNN models provide superior performance compared to the existing approaches for nuclei classification, segmentation, and detection tasks. The results are evaluated with different performance metrics including precision, recall, Dice Coefficient (DC), Means Squared Errors (MSE), F1-score, and overall accuracy. We have achieved around 3.4% and 4.5% better F-1 score for nuclei classification and detection tasks compared to recently published DCNN based method. In addition, R2U-Net shows around 92.15% testing accuracy in term of DC. These improved methods will help for pathological practices for better quantitative analysis of nuclei in Whole Slide Images(WSI) which ultimately will help for better understanding of different types of cancer in clinical workflow.

## 1.Introduction

People all around the work is suffering from different diseases including cancer, heart diseases, chronically diseases, Brain-related diseases including Alzheimer's, and diabetes. Recently study shows that one of the very famous company named "Pfizer" which has been conducting research for developing a drug for Alzheimer and Parkinson's diseases going to stop working on new drugs to fight Alzheimer's disease and Parkinson's disease due to hugely expensive and longtime. On the other hand, according to the recent study shows that on an average to discover a new drug it takes around 12 years to come to the market [1].

Medical imaging speed up the assessment process of almost every disease from lung cancer to heart disease. The automatic nucleus classification, segmentation, and detection algorithm can help to unlock the cure faster from the critical disease like cancer to the common cold. To identify the cell's nuclei is the starting point to analysis about 30 trillion cells contain a nucleus full of DNA of the human body. Identifying the cell accurately can help the researcher to observe how to react the cell with respect to the different treatments. As a result, researchers can understand the underlying biological process of cell-level analysis at clinical workflow. This solution can help to ensure the better treatment of patients and can accelerate the treatment for the patient and the drug discovery process. Therefore, the computational pathology and microscopy images play a big role in decision making for disease diagnosis, since these images able to provide a wide range of information for computer-aided diagnosis (CAD), which enables quantitative and qualitative analysis of these images with a very high throughput rate. Nowadays, the computational pathology becomes very popular in the field of medical imaging research which can greatly benefit pathologist and patient, therefore this field significantly get attention from both research community and the community from clinical practice [2,3]

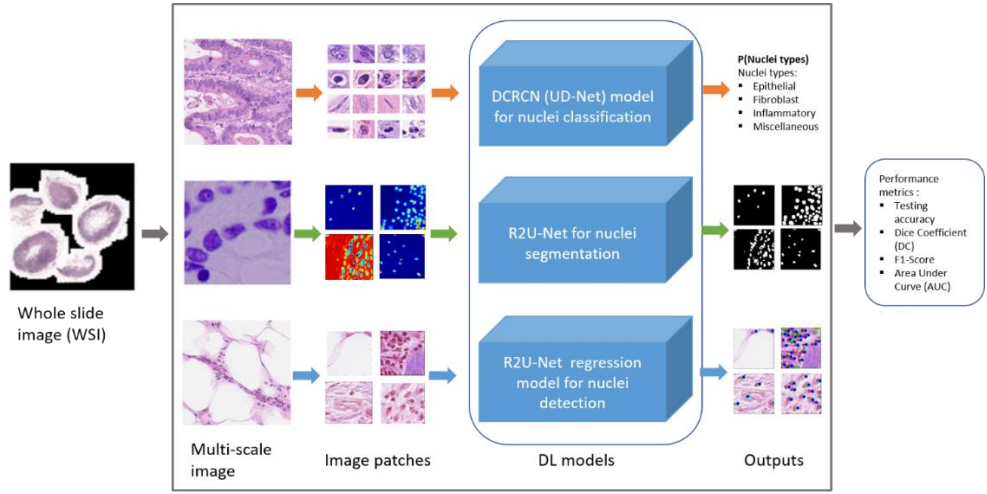

Figure 1: Overall proposed architecture: the microscopy WSI are acquired with different magnification factors, the patches are extracted from the multi-scale as required. Different DL model is applied for nuclei classification, segmentation, and detection. Eventually, the performance is evaluated with different performance metrics.

This computation approaches able to provides faster and efficient image analysis compare to the manual system for the researchers and clinician scientists which can release from difficult and repeated routine efforts [4]. Since the computational pathology and microscopic imaging is very challenging for manual image analysis, it might lead to large inter-observer variations [5]. On the other hand, CAD reduces the bias significantly and provide a characterization of diseases accurately [6]. Additionally, computational pathology gives a reproducible and rigorous measurement of pathological image features, can be used for clinical follow-up and helps to study personalized medicine and treatment which would significantly benefit patients.

As a prerequisite of clinical practice of CAD is nuclei classification, segmentation, and detection are considered as basically annotated image analysis method with DCNN. These techniques provide different quantitative analysis including cellular morphology, such as size, shape, color, texture, and other imagenomics. However, it is a very difficult task to achieve robust and accurate performance for mentioned three different challenging tasks for pathological imaging due to several reasons. First, the pathologically and microscopy images contain background clutter with the noise, artifacts (image are blur sometimes), low signal to noise ratio (SNR), and poor depth resolution, which is usually happened during image acquisition with devices. Second, it contains low contrast between the foreground and the background of the images. Third, the difficult issue is the variant of the size, shape, and intercellular intensity of nuclei or cell. Fourth, it can be observed very often that the nuclei or cells are partially overlapped with one another. Meanwhile, there are several methods have been proposed to tackle these issues with automatic nuclei classification, segmentation, and detection for pathological imaging.

In the last few years, there are several surveys have been conducted on different methods and summarized the CAD technologies in the field of biomedical imaging including computational pathology [7]. These reviews briefly discuss technique related to pre-processing, nuclei classification, segmentation, and detection, and post-processing. One of the recently published paper discusses several techniques related to data acquisition and ground truth generation, image analysis, recognition, detection, segmentation, and statistics in terms of survival analysis in [8]. Another review has conducted on different approaches related to feature extraction, predictive modeling, and visualization from WSI in [9]. A survey conducted on nuclei detection, segmentation, and classification on hematoxylin and eosin (H&E) and immunohistochemistry (IHC) stained histopathology images. Along with traditional image processing and computer vision-based approaches, there are several surveys have been on deep learning-based approaches for pathological image analysis. Due to available annotated data and huge computing power, the Deep Learning(DL) approached Convolutional Neural Network (CNN) providing state-of-the-art-accuracy on different computer vision problems [10]. First, in a classification task, the target is to identify the class probability from the input samples. From example: for binary breast cancer recognition problem the system defines class whether the input sample is in the category of benign or malignant. Second, most of the cases the semantics segmentation techniques are used for deep learning-based methods which describe the process of associating each pixel of an image with a class label. Another objective of this task is to define the proper contour of an object in the input image. Third, the DCNN based detection tasks, the objective is to identify the central or object rectangular coordinate of a certain object. Define the bounding box of an object is also the goal of this task. For example: in this implementation, identify the center pixel coordinate of nuclei from the input image. A recent study shows that the DL methods show a huge success in different modalities of medical imaging domain including mammographic mass classification, segmentation of lesions from neuroimaging, leak detection in airway tree, diabetic retinopathy, prostate segmentation, lung nodule detection, Breast cancer detection, x-ray imaging [11]. However, in this work, we have used applied three different improved DCNN models for nuclei classification, segmentation, and detection. The overall implementation diagram is shown in Figure 1. The contribution of this paper is summarized as follows:

- We have proposed an improved model named Densely Connected Recurrent Convolutional Network (DCRN) for nuclei classification.
- To generalize the R2U-Net model (UD-Net), this model is applied here for nuclei segmentation task in this implementation.
- The R2U-Net regression model is proposed for end-to-end nuclei detection from pathological images.
- The experiments have been conducted on three different publicly available datasets for nuclei classification, segmentation, and detection.
- The results show superior performance compared to existing machine learning and recently developed DL based approaches for nuclei classification, segmentation, and detection tasks.

The rest of the paper has been organized in the following way: Section II explains related works and Section III describes three different models. The database, results, and discussions are provided in Section IV. Conclusions and future directions are presented in Section V.

## 2. Related works

Automatic nuclei classification, segmentation, and detection is a prerequisite for various quantitative and qualitative analysis in computational pathology. The morphological features compute for different disease including routine colon cancer, breast cancer, drug development and many more. In the last few years, there are different DCNN approaches has been proposed and successfully applied on medical image analysis problems and shows superior performance on different benchmarks dataset for classification, segmentation, and detection task [1]. There are several types of research ongoing in the field of digital pathology and trying to improve the performance due to the complex nature of images. However, we have considered the nuclei classification, segmentation, and detection have been treated as a separate problem. The related works on traditional machine learning and deep learning-based nuclei classification, segmentation, and detection is as follow.

**Classification:** Nuclei classification can be used to various histopathology related applications. In the past, features including shape, texture, and size of nuclei are considered for nuclear pleomorphism grading in breast cancer images [12]. Malon et al. have applied CNN for classifying the mitotic and non-mitotic cells using color, shape and texture information in [13]. Cancerous nuclei are classified lymphocyte or stromal based on morphological features in H&E stained for breast cancer images and the method requires an accurate segmentation of tissue from the input samples which is explained in [14]. Another method with AdaBoost classifier where intensity, morphological and texture feature are used and the main focused of that work was on nuclei segmentation with classification approach [15]. However, recent studies have shown the deep learning-based approaches produce promising classification accuracy on large-scale pathological images. In 2014, Wang et al. used hand-crafted featured and applied cascaded ensemble CNN for detecting mitotic cells and achieved promising improvement result for nuclei classification task in [16]. Another deep learning based approach is proposed for cell classification and compared against a method with a bag of features and canonical representations in [17]. In 2017, the histopathological image classification approach is proposed and applied support vector machine (SVM), AdaBoost, and DCNN. The experiment is conducted on four different H&E stained image datasets namely prostate, breast, renal clear cell, and renal papillary cell cancer dataset. The results demonstrate that Color-Encoder deep network achieves the best performance out of nine individual methods and they achieved around 91.2% testing accuracy in term of F1-score as highest testing accuracy [18].

**Segmentation:** a novel contour based "minimum-model" cell detection and segmentation approaches are proposed in 2012. That method uses minimal a priori information and detects contours independent of their shape and achieved promising segmentation results in [19]. Nuclei membrane segmentation with the CNN model is proposed from microscopic images in [20]. Ronneberger et al. proposed a CNN based approach called U-Net for general medical image segmentation in [21]. In addition, the U-Net has been applied range of segmentation problem including Nuclei segmentation. A learning-based framework for robust and automatics nucleus segmentation with shape preservation in pathological image, the CNN model generates the probability hit maps, on which an iterative region merging technique is applied for shape identification. In addition, a Nobel segmentation approach was exploited to separated individual nuclei combining a robust selection-based shares shape model and a local repulsive deformable model which have tested in several scenarios for pathological image segmentation and shows state-of-the-art performance against existing approaches till 2016 [22]. A very simple CNN model-based nuclei segmentation approach is proposed in 2017 which are named CNN2, and CNN3 models with respect to the number of output classes. For the two-class model, the network is used to classify pixel for inside and outside of the nuclei region respectively. On the other hand, the model for three classes, they were used for classifying pixels belong to inside, an output side, and the boundary of the nuclei regions in [23]. In 2017, Ho, D.J. et al. have proposed a fully 3D nuclei segmentation method using 3D CNN [24]. In 2018, another very promising deep learning-based one-step contour aware nuclei segmentation approach is proposed where a fully convolutional neural network is applied to segment the nuclei with their boundaries simultaneously [25].

A 3D Convolutional Network is used for joining cell Nuclei detection and simultaneously segmentation in microscopic images. The model is tested on two different datasets and achieved state-of-the-art accuracy in detection and segmentation tasks [26]. However, for general medical image segmentation, an improved version of U-Net deep learning model has proposed in 2018 where recurrent residual modules are incorporated in U-Net instead of feedforward convolutional layers. The model was evaluated on different modalities of medical imaging including retinal blood vessel segmentation, skin cancer segmentation, and lung segmentation. The experimental results are compared against U-Net, and SegNet and show superior testing performance [27]. To generalized of R2U-Net model, we have used R2U-Net model for end-to-end nuclei segmentation in this implementation. Along with that, the nuclei classification and detection methods are included as an extension of the primary results were published in 2018 [28].

**Detection:** Nowadays, there are two different methods are mainly used for nuclei detection. First, detection-based counting, which requires a prior detection or segmentation that discussed in [29]. On the other hand, density estimation-based method is used for nuclei detection without using segmentation which is explained in [30]. A framework with supervised max-pooling CNN is trained to detection cell pixels region which is preselected with Support Vector Machine (SVM). The method has shown outperformance compare to hand-crafted features-based approaches in [31]. For nuclei detection, a stacked sparse autoencoder based approach is used for non-nuclei and nuclei region detection with unsupervised fusion where a Softmax classifier is used in [32]. A CNN based regression model is used for nuclei detection and counting where a fully convolutional neural regression network model is used and able to density map for an input image of arbitrary size. They were used a patch-based method instead of end-to-end image method which is explained in [33]. However, we have proposed R2U-Net based regression model for end-to-end nuclei detection in this implementation.

## 3. Proposed Deep CNN models

### 3.1 Densely Connected Recurrent Convolutional Network (DCRN)

According to the basic structure of Densely Connected Networks (DCN), the outputs from the prior layers are used as input for the subsequent layers. This architecture ensures the reuse the features inside the model, therefore it provides better performance on different computer vision tasks which in empirically investigated on different datasets in [34]. However, in this implementation, we have proposed an improved version of DCN which is named DCRN in short which is used for nuclei classification. The UD-Net is the building block of several Recurrent Connected Convolutional (DCRC) blocks and transition blocks. The pictorial representation of Densely Connected Recurrent Convolutional (DCRC) block is shown in Figure 2.

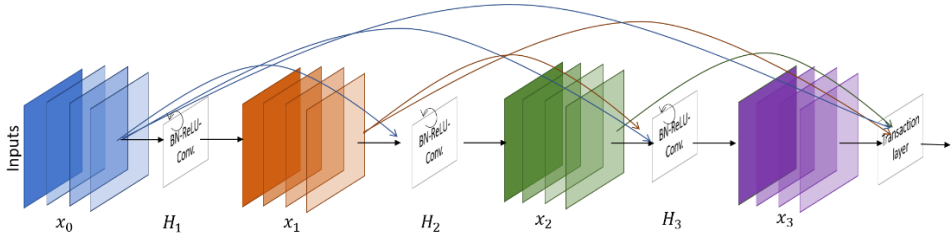

Figure 2: Densely Connected Recurrent Convolutional (DCRC) block.

According to the basic mathematical model of DenseNet which has explained in [34], the $l^{th}$ layer receive all the feature maps $(x_0, x_1, x_2 \cdots x_{l-1})$ from the previous layers as input:

$$x_l = H_l([x_0, x_1, x_2 \cdots x_{l-1}]) \qquad (3)$$

where $[x_0, x_1, x_2 \cdots x_{l-1}]$ is the concatenated features from $0, \cdots\cdots, l-1$ layers and $H_l(\cdot)$ is a single tensor. Let's consider the $H_l(\cdot)$ input sample from $l^{th}$ DCRN block and contains $0, \cdots\cdots, F-1$ feature maps which are feed in the recurrent convolutional layers according to the method has proposed in [194,196]. This convolutional layer performs three consecutive operations which include Batch Normalization (BN), followed by ReLU and a $3 \times 3$ convolution (conv). Let's consider a center pixel of a patch located at $(i,j)$ in an input sample on the $k^{th}$ feature of $H_{(l,k)}(\cdot)$. Additionally, let's assume the output of the network is $H_{lk}(t)$ for $l^{th}$ layer and $k^{th}$ feature maps atthe the time step t. The output can be expressed as follows:

$$H_{lk}(t) = \left(w_{(l,k)}^f\right)^T * H_{(l,k)}^{f(i,j)}(t) + \left(w_{(l,k)}^r\right)^T * H_{(l,k)}^{r(i,j)}(t-1) + b_{(l,k)} \qquad (1)$$

Here $\mathbf{H}_{(l,k)}^{f(i,j)}(t)$ and $\mathbf{H}_{(l,k)}^{r(i,j)}(t-1)$ are the inputs to the standard convolution layers and the $l^{th}$ recurrent convolution layers respectively. The $\mathbf{w}_{(l,k)}^f$ and $\mathbf{w}_{(l,k)}^r$ values are the weights of the standard convolutional layer and the recurrent convolutional layers of $l^{th}$ layer and $k^{th}$ feature map respectively, and $\mathbf{b}_{(l,k)}$ is the bias. The recurrent convolution operations are performed with respect to $\mathbf{t}$ [35,37]. The pictorial representation of convolutional operation for $\mathbf{t=2}$ is shown in Figure 3.

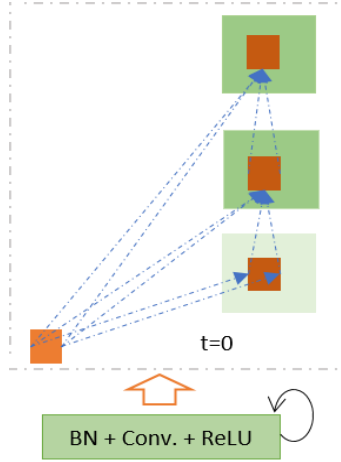

Figure 3: Unfolded recurrent convolutional units for t = 2.

In the transition block, $\mathbf{1 \times 1}$ convolutional operations are performed with BN followed by $\mathbf{2 \times 2}$ average pooling layer. The DenseNet model is consisted of several dense blocks with feedforward convolutional layers and transition blocks whereas the DCRN uses samethe number DCRC units and transition blocks. For both model, we have used 4 blocks, 3 layers per block, and growthe th rate is 5 in this implementation and the model details are given in Table 1.

**3.2 R2U-Net**

we have applied the R2U-Net model for nuclei segmentation from microscopic images in [27]. The R2U-Net has been constructed with U-Net [21], Recurrent Convolutional Neural Networks (RCNN) [36], and Residual Network (Reset) [38]. The entire R2U-Net model is provided in Figure 104. This model consists of two main units which are encoding unit (shown in green) and a decoding unit (shown in blue). In both units, the recurrent residual convolutional operations are performed in the convolutional blocks.

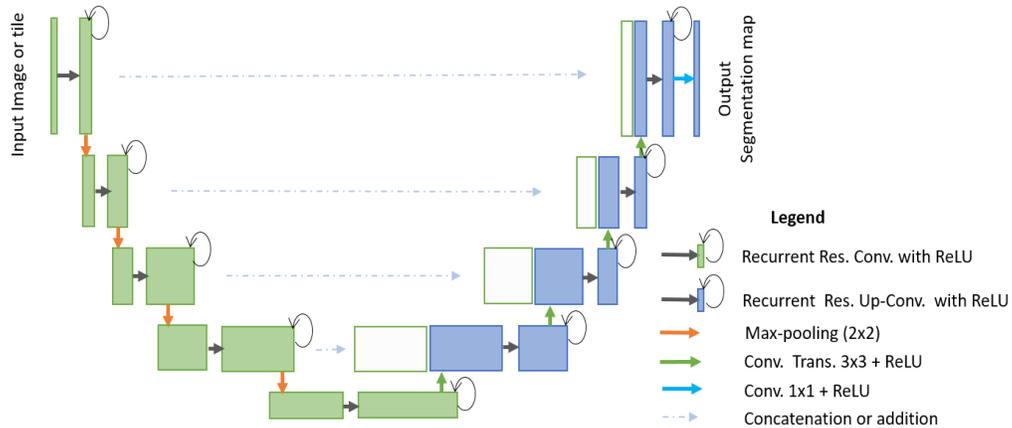

Figure 4: The end-to-end Nuclei segmentation method with R2U-Net model: green part refers the encoding unit, and blue part stands for decoding units. The features are concatenated from encoding units to the decoding units.

The conceptual diagram of the recurrent residual unit is shown in Figure 5. The recurrent operation is performed with respect to different time steps, which is shown in Figure 3. For the recurrent convolutional unit, t = 2, which means one forward convolution layer and two recurrent layers are used in this convolutional unit. The feature maps from the encoding unit are concatenated with the feature maps from decoding units. The softmax layer is used at the end of the model to calculate the pixel label class probability. For further details about R2U-Net, please see [27]. The model details and number of feature maps for this implementation are shown in Table 1.

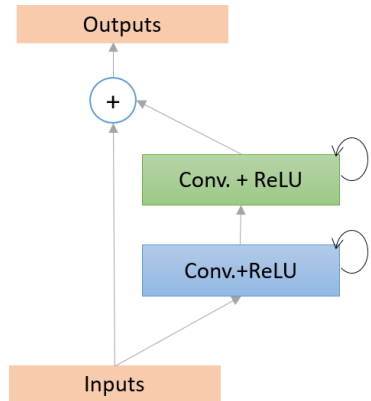

Figure 5: The recurrent residual unit (RRU) for R2U-Net.

**3.3 Regression model with R2U-Net**

In general, for cell detection and counting problem, the gourd truth is created with a single pixel annotation where the individual dot represents a cell. For example: the dataset, we have used in this implementation contains around at least five to five hundred nuclei with center pixel of the cell in input samples. For training with a regression model, each dot is represented with a Gaussian density. In case of the regression model, we have applied R2U-Net model to estimate the Gaussian densities from the input samples instead of computing the class or pixel level probability which is considered for DL based classification and segmentation model respectively. This model is named the University of Dayton Network shortly "UD-Net". For each input sample, a density surface $D(x)$ is generated with superposition of these Gaussian. The objective is to regress this density surface for the corresponding input cell image $I(x)$. The goal is achieved with R2U-Net model with the mean squared errors loss between the output heat maps and the target Gaussian density surface which is the ultimate loss function for the regression problem. However, in the inference phase, for the given input cell image $I(x)$, the model R2U-Net computes the density heat maps $D(x)$. A CNN-VGG architecture-based regression model is explained in [42]. However, in this work, we have proposed regression model with R2U-Net. The details on R2U-Net is described in the previous section, the network architecture of the regression model and the number of network parameters are shown in Table 1.

Table 1: The model configuration and a number of network parameters utilize in this implementation.

| Model | Tasks | t | Network architectures | Number of parameters (million) |
|---|---|---|---|---|
| DenseNet | Classification | - | Blocks #4, layers#3, and growth rate # 5 | 0.582 |
| DRCN | Classification | 2 | Blocks #4, layers#3, and growth rate # 5 | 0.582 |
| R2U-Net | Segmentation | 2 | 1→ 16→32→64→28→64→32→16→1 | 0.845 |
| UD-Net | Detection | 3 | 1→16→32→64→128→64 →32→16→1 | 1.038 |

Network architectures: We have used similar architecture for DenseNet and DRCN models only difference between these two network models are the recurrent connectivity in the convolutional unit. In the case of DenseNet model, the two feedforward convolutional layers are used whereas for DRCN, we have two recurrent convolutional layers. For segmentation, we have used R2U-Net model with 0.84M the network parameters which is implemented for $t=2$. However, we have applied DRCNN model with $t=3$ which increases the number of network parameters (1.038M).

## 4. Experiments and results

To demonstrate the performance of the DCRN, R2U-Net, and R2U-Net based regression (UD-Net) models, we have tested them for the nuclei classification, segmentation, and detection tasks. The dataset for classification and detection tasks are taken from [39] and segmentation dataset is downloaded from the 2018 Data Science Bowl Grand Challenge [45]. A discussion of this dataset is provided in the following sections. For this implementation, the Keras [46] and TensorFlow [47] frameworks were used on a single GPU machine with 56G of RAM and a NVIDIA GEFORCE GTX-980 Ti.

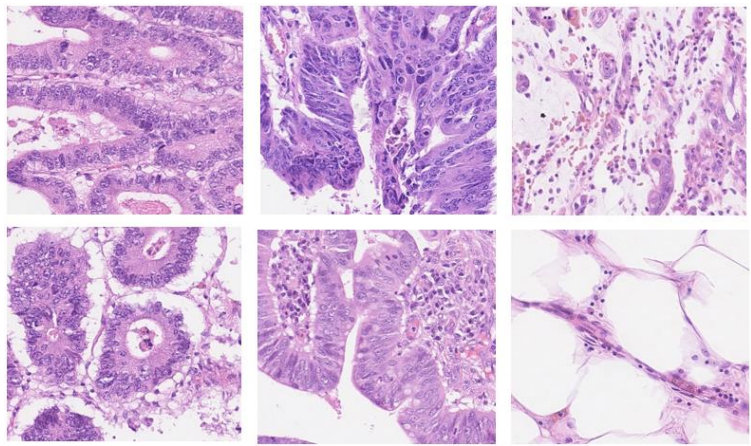

Figure 6: Example images from the dataset for Nuclei classification

## 4.1 Dataset for nuclei classification

This dataset contains 200 annotated samples for classification and detection tasks, where the total number of 100 samples are utilized for classification task and remaining 100 samples are used for detection task respectively. The actual sample size is 500x500 pixels. Some of the randomly selected database samples for nuclei classification are shown in Figure 6. In this implementation, we have selected 200 random patches from each sample, the selected patches size is 32x32 pixels. This dataset has four different classes of routine colon cancer including Epithelial, Fibroblast, Inflammatory, and miscellaneous. For the classification task, we have had total 20,000 samples [39]. The example patches with respective classes are given in Figure 7.

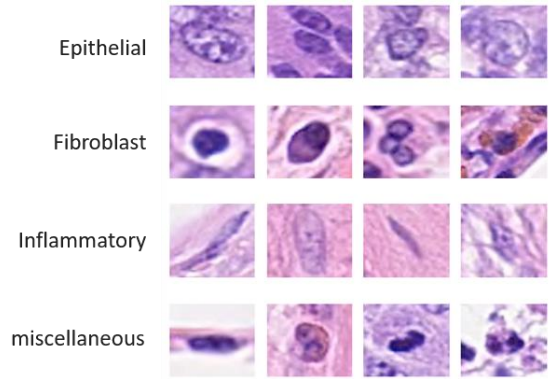

Figure 7: Image patches for four different routine colon cancer

## 4.2 Dataset for nuclei segmentation

In 2018, Data Science Bowl has launched a competition with a mission to create an effective algorithm for automatic nucleus detection and segmentation. This work utilizes the dataset from 2018 Data Science Bowl grand challenges [45], which contains 735 images in total. The size of the database sample is 256×256 pixels. From the total image, 650 images and corresponding with pixel-level annotation masks are considered for training and remaining 65 samples are unlabeled to testing. From the training set, 80% of samples are used for training and the remaining 20% are used for validation during training. The number of training and validation samples are 536 and 134 respectively. The randomly selected some samples from the training set are shown in Figure 8. Figure 9 shows the normalized training samples in the first rows and corresponding label in the second row.

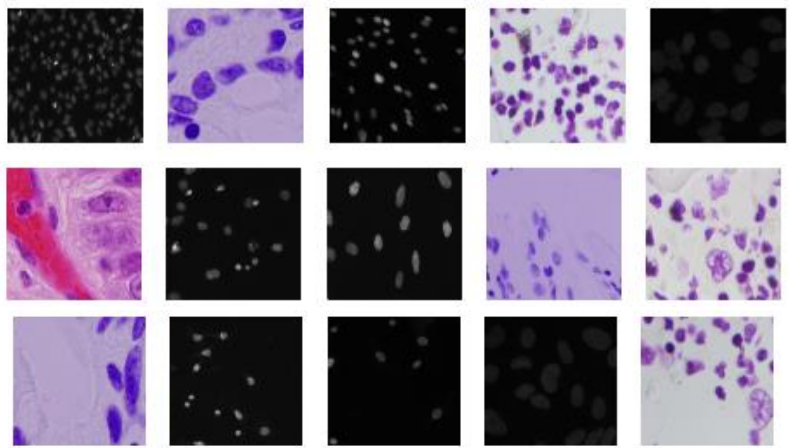

Figure 8: Randomly selected images from the dataset for segmentation.

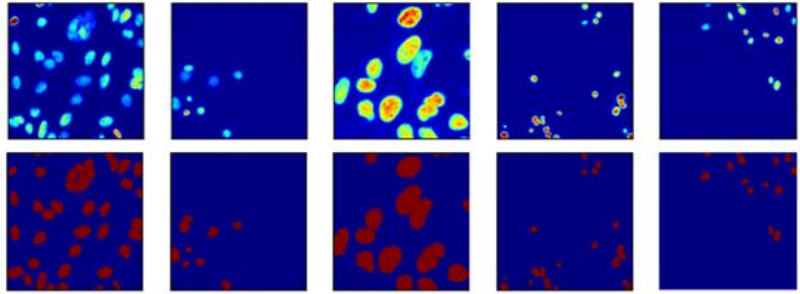

Figure 9: Example images with labels: first normalized samples and the second row shows the label of the corresponding images.

### 4.3 Database for Nuclei Detection

The database contains 100 samples and 100 masks with single pixel annotation [39]. The original size of the database sample is 500×500. For detection task, the nuclei cells are usually annotated with a single dot which is the center pixel of nuclei. For better understanding, some of the randomly selected samples and corresponding single pixel annotated masks are shown in Figure 10.

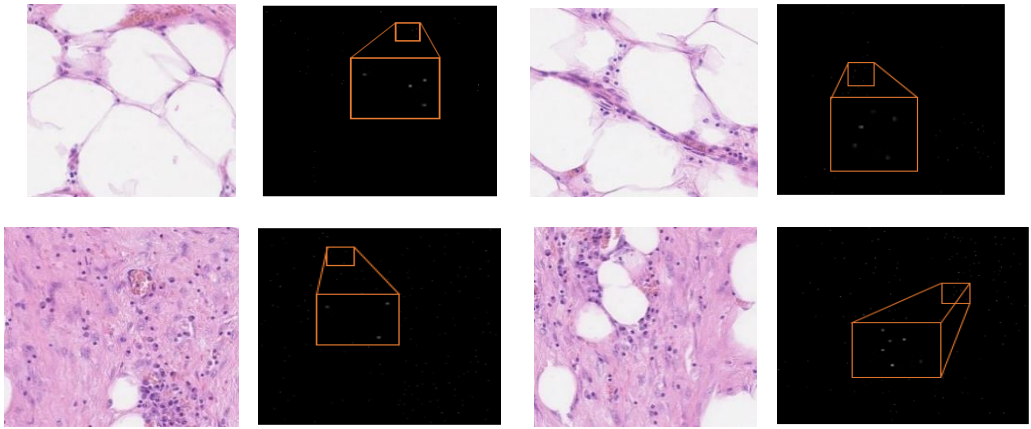

Figure 10: Example image with corresponding single pixel annotated masks for nuclei detection.

In this implementation, we have selected the non-overlapping patches (96×96 pixels) from original input samples and corresponding masks. We have extracted total 4,392 non-overlapping patches. From these patches, around 80% patches are used for training and the remaining 20% samples are used for testing.

### 4.4 Training methods

To train models for the classification task, we have applied DenseNet and DCRN with the same architecture and network parameters. In both cases, we have used stochastic gradient descent(SGD) optimization method with learning rate of 0.001, weight decay $1\times10^{-4}$, momentum 0.9 and cross entropy loss. The entire model is trained for 100 epochs with batch size 32. For the segmentation task, we have applied the Dice Coefficient (DC) and Mean Squared Error (MSE) loss function in this implementation. The DC is expressed in equation following Eq. (1), where GT refers to the ground truth and SR refers the segmentation result.

$$DC = 2 \frac{|GT \cap SR|}{|GT|+|SR|} \tag{1}$$

Another metric used to evaluate the performance of the segmentation algorithm is the MSE as defined in equation (2). In this case, Y represents desired outputs and $\hat{Y}$ represents the predicted outputs. For an input sample with height $h$ and width $w$ and $n=h \times w$.

$$MSE = \frac{1}{n} \sum_{i=1}^{n}(Y_i - \hat{Y}_i)^2 \tag{2}$$

We considered 250 epochs and used Adam optimizer with a learning rate of $2 \times 10^{-4}$ and the batch size is 16. Finally, for detection task with UD-Net regression model, we have used Adam optimizer with learning rate $2 \times e$-4 and means squared errors (MSE). The model is trained with 500 epochs and batch size 64.

**4.5 Results and discussion**

*4.5.1 Nuclei classification*

We have tested models with a completely separated dataset from training dataset and achieved 90.86% and 91.14% testing accuracy with DenseNet and DCRN models respectively. The experimental results and comparison against the existing approaches are shown in Table 2. This table demonstrates that DRCN shows superior performance compared to existing methods.

Table 2: Nuclei classification accuracy and compared against other methods

| Methods | Average F1-score | AUC | Accuracy |
|---|---|---|---|
| CRImage [40] | 0.488 | 0.684 | - |
| Super-pixel descriptor [40] | 0.687 | 0.853 | - |
| SoftMax CNN + SSPP [39] | 0.748 | 0.893 | - |
| SoftMax CNN + NEP [39] | 0.784 | 0.917 | - |
| DenseNet [34] | 0.8121 | 0.958 | 0.9086 |
| Proposed (DRCN) | 0.8180 | 0.9615 | 0.9114 |

The area under Receiver Operating Characteristic (ROC) curve for both DenseNet and DRCN is shown in Figure 11. From this figure, it can be clearly seen that DRCN provides show higher Area Under Curve (AUC) with similar network architecture and a same number of network parameters which clearly demonstrates the robustness of our proposed model over DenseNet for nuclei classification.

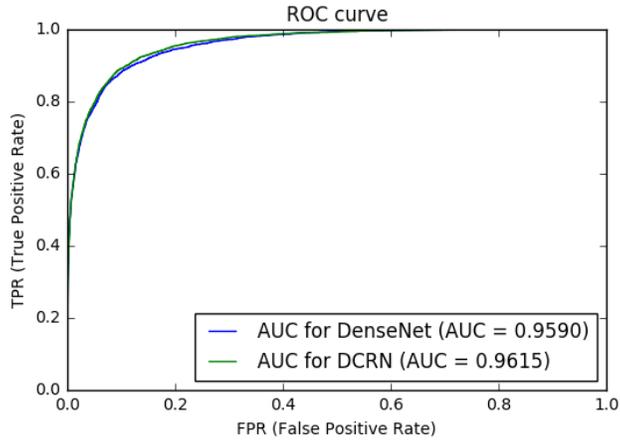

Figure 11: Area under the ROC curve for DenseNet and DCRN models.

### 4.5.2 Nuclei segmentation

In this implementation, we used a simple R2U-Net model where only 0.84 million network parameters were utilized. The network architecture along with a number of feature parameters is shown in Table 1. We considered the DC and MSE for observing training progress and performance in the training and testing phases. Figure 12(a) shows the DC and MSE with respect to the number of epochs during training. The validation DC and MSE are shown in Figure 12(b). From these figures, it can be observed that the model converged after 100 epochs, the training and evaluation continued until 250 epochs were completed to ensure optimum convergence. From this experiment, we achieved approximately 92.15% testing accuracy for nuclei segmentation with the R2U-Net model.

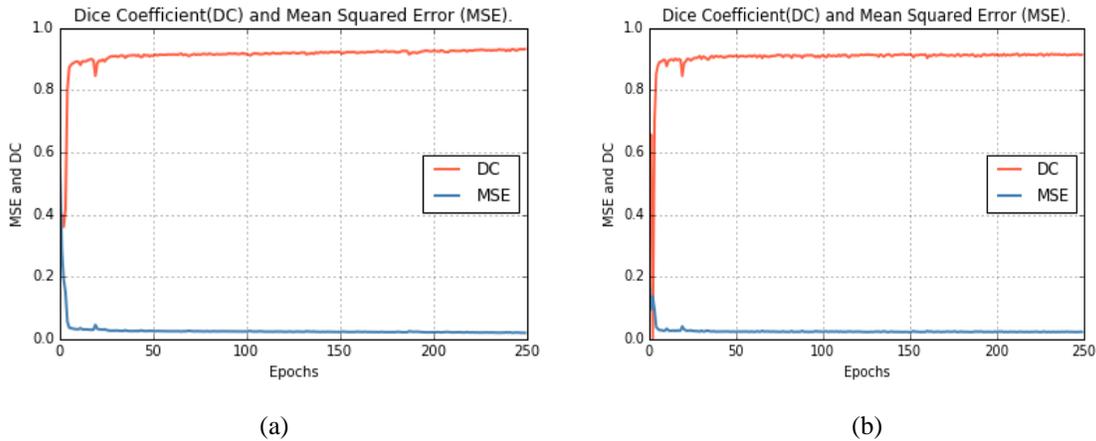

(a)             (b)

Figure 12: Training and validation accuracy in term of the Dice Coefficient (DC) on the left and Mean Squared Error (MSE) on the right for 250 epochs.

**Qualitative results:** Figure 13 shows some example output samples when using the R2U-Net model for nuclei segmentation where the first and third rows show the input samples. Likewise, the second and fourth rows represent the corresponding network outputs. Based on these results, our proposed segmentation model provides promising segmentation outputs during the testing phase. In addition, if we observe the input samples in the first and second columns of the first row, there is a strong separation between nuclei and similar objects of less interest. Similar behavior is displayed in the input sample in the first row of the fourth column. The input shown in the third row of the fourth column contains a complex background. However, in all cases, the model shows almost similar segmentation outputs with respect to the desired outputs. This result clearly demonstrates the robustness of R2U-Net for nuclei segmentation from pathological images.

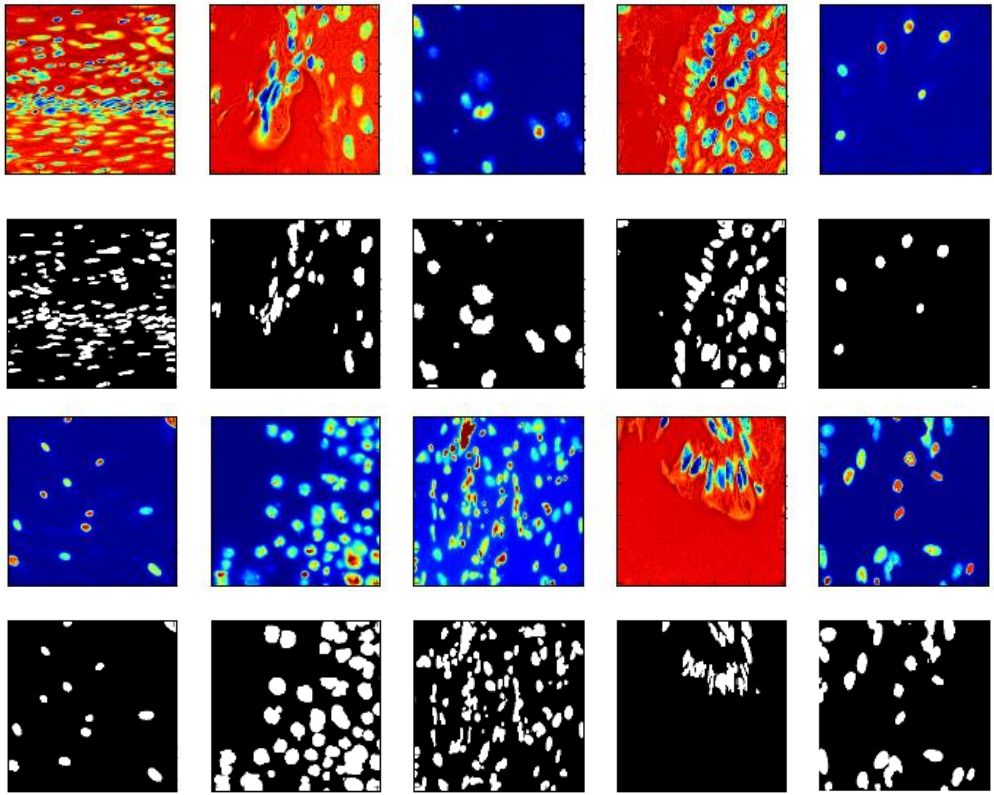

Figure 13: Qualitative experimental outputs in the testing phase with R2U-Net: first and third rows show the testing inputs and second and fourth rows show outputs samples.

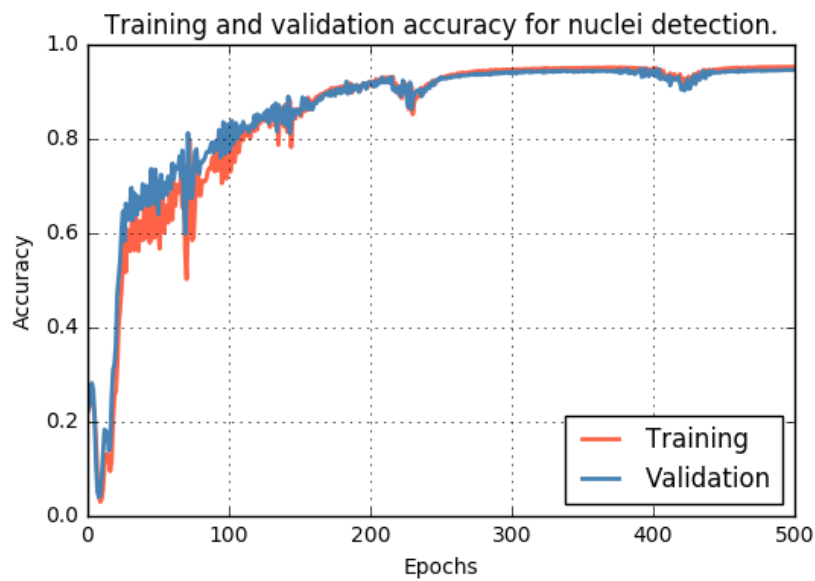

Figure 14: Training and validation accuracy of R2U-Net based regression model for nuclei detection

### 4.5.3 Nuclei detection

In this experiment, we have considered a patch-based approach, the experimental results are shown in Table 3. The recently published paper has reported 0.802 F1-score as the highest testing accuracy for nuclei detection in 2016 whereas our proposed model shows 0.8318 for nuclei detection task which is around 2.98% higher performance over the recently published SC-CNN model in [39].

Table 3: Nuclei detection accuracy and compared against other methods

| Methods | Precision | Recall | Average F1-score |
| --- | --- | --- | --- |
| CRImage [40] | 0.657 | 0.461 | 0.542 |
| CNN[40] | 0.783 | 0.804 | 0.793 |
| SSAE [43] | 0.617 | 0.644 | 0.630 |
| LIPSyM [43] | 0.725 | 0.517 | 0.604 |
| SC-CNN [39] (M=1) | 0.758 | 0.827 | 0.791 |
| SC-CNN [39] (M=2) | 0.781 | 0.823 | 0.802 |
| Proposed (UD-Net) | 0.821 | 0.842 | 0.832 |

The quantitative results for nuclei detection are shown in Figure 15. The result shows promising detection accuracy for input patches. The first column shows the input patches, the second column shows the inputs label masks, third rows represents the model outputs, and the fourth column shows the final outputs with blue and green dots. Here the blue dot indicates the ground truth and the green dot represents the model outputs. The quantitative results clearly demonstrate that the UD-Net able to detect the nuclei pretty accurately.

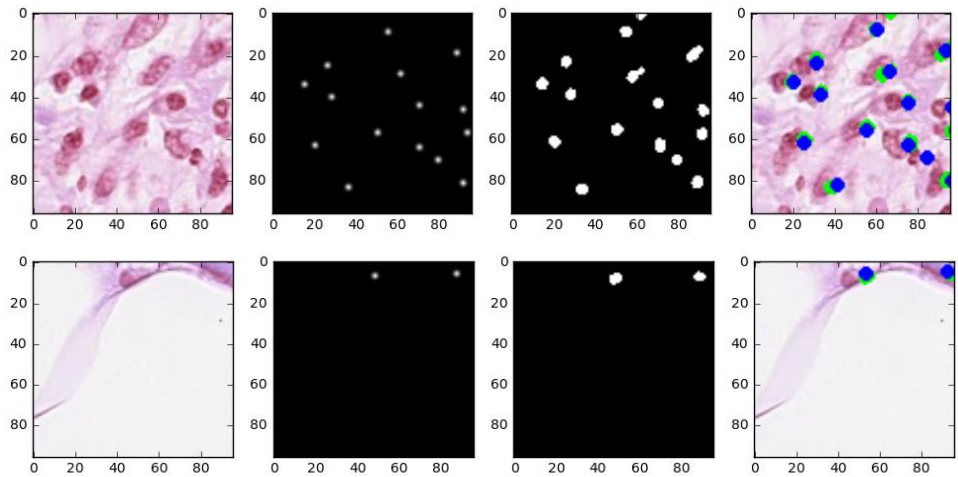

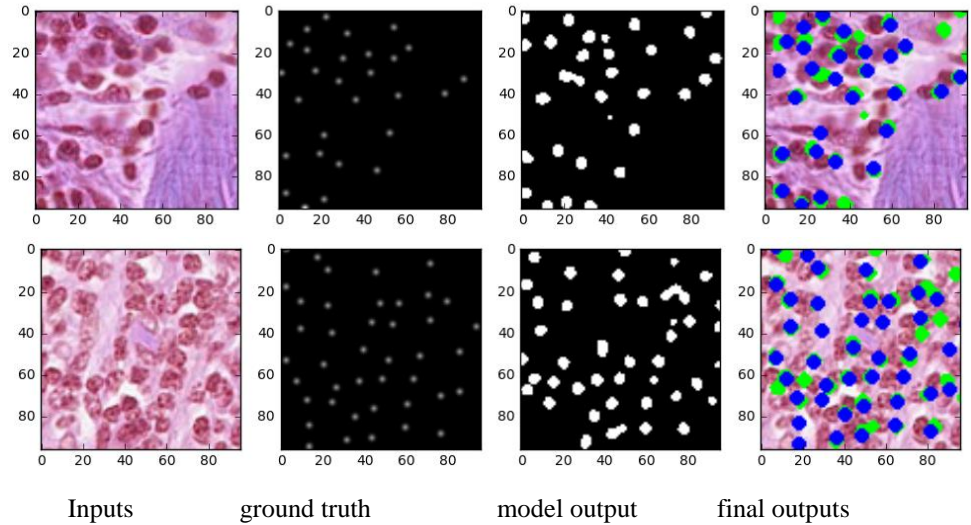

Inputs        ground truth        model output        final outputs

Figure 15: Nuclei detection outputs with inputs, ground truth, network outputs, and final outputs with a blue and green dot. The blue dot represents the center pixel of ground truth and green dot shows center pixels of the network outputs.

Entire Images based outputs: after generating the patch-based outputs, we have merged all the patches together to generate the entire input image and corresponding outputs. The Figure 16, shows the output for entire input image where the first column shows the inputs, the second column is for ground truth Gaussian density surface, the third column represents the model outputs and the fourth column shows the final outputs with a blue and green dot for each nucleus. The blue dot represents the ground truth and the green dot represents the model's outputs.

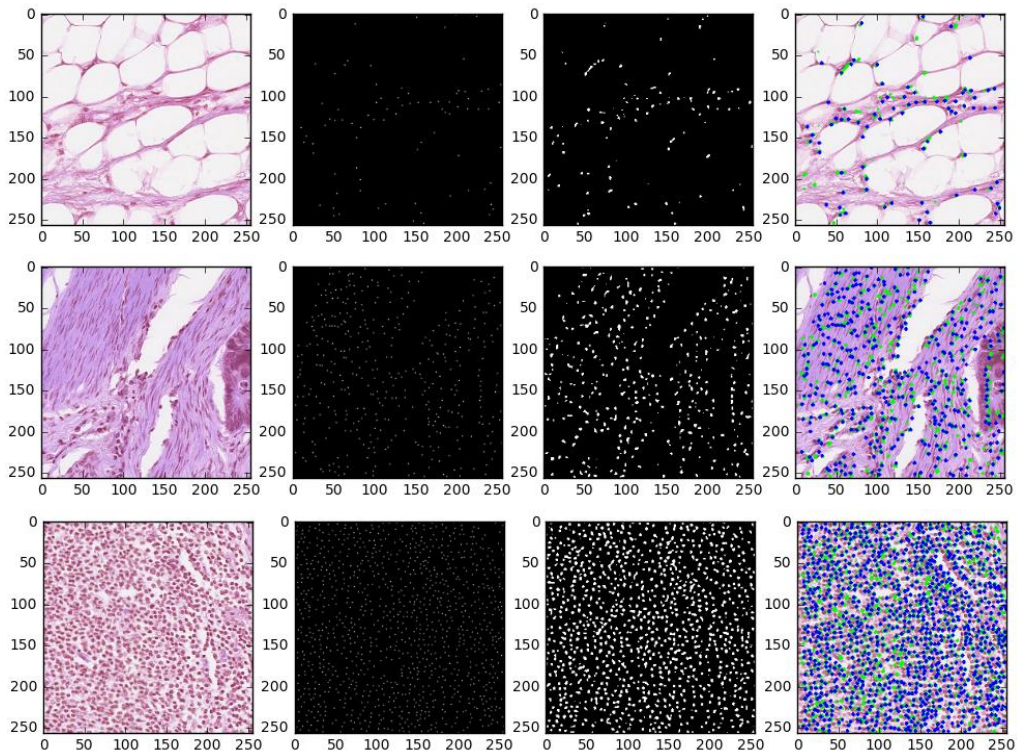

Figure 16: Nuclei detection outputs for entire samples which are generated from output patches.

**4.6 Analysis**

We have solved three important tasks for computational pathology: nuclei classification, segmentation, and detection. In classification, we have applied DenseNet and improved version of DenseNet is named DCRN. The DenseNet provides 0.8121 and 0.958 performance in term of F1-score and AUC whereas the proposed DCRN provides around 0.8180 and 0.9615 for F1-score and AUC. The DCRN provides around 3.4% and 4.45% better performance for F1-score and AUC against recently published results in [39]. In addition, our proposed method shows 91.17% testing accuracy which around 0.3% better compared to DenseNet. Second, we have used very simple R2U-Net where only 0.84 million (M) network parameters for nuclei segmentation, we have achieved 92.15% testing accuracy on the publicly available dataset. Third, R2U-Net based regression model is used for Nuclei detection task and achieved 82.17%, 84.23% and 83.2% for precision, recall, and F1-score respectively. Overall, our models provide superior performance for three tasks. The testing time per sample for classification, segmentation, and detection tasks are shown in Table 4.

Table 4: Computational testing time of proposed DCRN, R2U-Net, and UD-Net models in second.

| Model | Tasks | Computational time/epoch (in sec.) |
| --- | --- | --- |
| DCRN | Classification | 0.0017 |
| R2U-Net | Segmentation | 0.40239 |
| R2U-Net regression model | Detection | 3.19906 |

**5. Conclusion**

In this study, we have proposed three different models including Densely Connected Recurrent Convolutional Networks (DCRN), Recurrent Residual U-Net (R2U-Net), and R2U-Net based regression (UD-Net) model for nuclei classification, segmentation, and detection tasks respectively. These models have evaluated on publicly available three different datasets. We have achieved 91.14% testing accuracy with DCRN for nuclei classification task and achieved 3.4% and 4.45% higher than the average F1-score and AUC compare to recently published DL based method a softmax Convolutional Neural Networks(CNN) and neighboring ensemble predictor (NEP) which is called softmax CNN+NEP. The proposed R2U-Net model is applied for Nuclei segmentation task and shows 92.15% testing accuracy. For detection task, we have achieved 83.2% testing accuracy in term of F1-scope with R2U-Net regression model shows around 3% better F1-score compared to the existing methods. In the future, we would like to explore and evaluate these models on other datasets in the field of computational pathology.